\documentclass[
twocolumn,
]{ceurart}

\usepackage{listings}
\usepackage{soul}
\usepackage{url}
\usepackage[small]{caption}
\usepackage{graphicx}
\usepackage{amsmath}
\usepackage{amsthm}
\usepackage{booktabs}
\usepackage{algorithm}
\usepackage{algpseudocode}

\usepackage{tikz}
\usepackage{tikzscale}
\usepackage{comment}
\usepackage{caption}
\usepackage{subcaption}
\usepackage{pgfplots}
\usepgfplotslibrary{groupplots}
\usetikzlibrary{positioning}

\begin{document}

\copyrightyear{2023}
\copyrightclause{Copyright for this paper by its authors.
  Use permitted under Creative Commons License Attribution 4.0
  International (CC BY 4.0).}

\conference{KDH@IJCAI'23: Knowledge Discovery From Healthcare Data,
  August 20, 2023, Macao, S.A.R}

\title{Anderson Acceleration For Bioinformatics-Based Machine Learning}

\author[1]{Sarwan Ali}[%
email=sali85@student.gsu.edu,
]
\fnmark[1]
\address[1]{Georgia State University, Atlanta, USA}

\author[1]{Prakash Chourasia}[%
email=pchourasia1@student.gsu.edu,
]
\fnmark[1]
\author[1]{Murray Patterson}[%
email=mpatterson30@gsu.edu,
]
\cormark[1]

\cortext[1]{Corresponding author.}
\fntext[1]{These authors contributed equally.}

\begin{abstract}
  Anderson acceleration (AA) is a well-known method for accelerating the convergence of iterative algorithms with applications in various fields, including deep learning and optimization. Despite its popularity in these areas, the effectiveness of AA in classical machine learning classifiers has not been thoroughly studied. Tabular data, in particular, presents a unique challenge for deep learning models, and classical machine learning models are known to perform better in these scenarios. However, the convergence analysis of these models has received limited attention. To address this gap in research, we implement a support vector machine (SVM) classifier variant incorporating AA to speed up convergence. We evaluate the performance of our SVM with and without Anderson acceleration on several datasets from the biology domain and demonstrate that the use of AA significantly improves convergence and reduces the training loss as the number of iterations increases. Our findings provide a promising perspective on the potential of Anderson acceleration in training simple machine learning classifiers and underscore the importance of further research in this area. By showing the effectiveness of AA in this setting, we aim to inspire more studies that explore the applications of AA in classical machine learning.
\end{abstract}


\maketitle

\section{Introduction}

Anderson acceleration is a method that can be used to enhance the convergence of gradient descent algorithms. Based on the difference between the current and prior weight vectors, a correction term is added to the weight vector updates at each iteration. When the gradients are changing quickly, or the optimization landscape is very non-convex, this correction term can aid in reducing oscillations and speeding convergence. Consider the optimization issue as a trajectory in the weight space, where the weight vector reflects the position at each iteration, to appreciate this concept better. Without Anderson acceleration, the gradients at each location alone control the trajectory of the optimization process. While Anderson acceleration can smooth out the trajectory and minimize oscillations, the trajectory is also affected by the difference between the current and prior weight vectors.

Solving the convex problem in finding gradient descent is a typical problem in optimization. Newton's methods use the inverse Hessian matrix~\cite{goldfarb2020practical} to accelerate gradient descent, and they are successful in achieving a faster rate of convergence compared to gradient descent or accelerated gradient descent, but it is very expensive.
By utilizing knowledge of the curvature of the loss function landscape and quasi-Newton algorithms~\cite{tong2021asynchronous} that compute a low-rank approximation of the Hessian, it is feasible to accelerate the training of machine learning(ML) models. The approximate replacement matrix for the Hessian Inverse can be found using quasi-Newton methods, described in detail by authors in ~\cite{shanno1970conditioning}. The alternative option is to use the existing data that is already available for the fixed point approach, such as natural gradient descent (NGD) ~\cite{naganuma2019performance} and extrapolation methods for vector sequences~\cite{brezinski2022shanks}.
Approximation-based approaches such as popular techniques, including momentum acceleration methods ~\cite{qian1999momentum}, and stochastic Newton methods ~\cite{tong2021asynchronous}, can be used to calculate the approximate replacement matrix. However, as dimensionality and complexity rise, the benefits of these methods are outweighed by the computational expense. Due to the increased dimensionality and complexity of the data, evaluating the training loss function is getting increasingly expensive; consequently, fixed-point~\cite{walker2011anderson} approaches are more suited for hastening the training of ML models.
  
The machine learning (ML) community has become interested in extrapolation techniques like Anderson acceleration (AA)~\cite{brezinski2022shanks}. These techniques have been demonstrated to speed up the training of sequential deep learning models (DL)~\cite{shi2019regularized}.
Since the optimization landscape for deep learning models can be more complicated to improve than simpler models, Anderson acceleration is typically more successful when used in optimization methods for deep learning~\cite{keskar2016large}. Although it might not offer as big of a gain in terms of convergence speed compared to more complex models, Anderson acceleration may still be effective for optimizing simpler classical ML models. Ultimately, the precise characteristics of the optimization problem being solved will determine how effective Anderson acceleration is in each given scenario.

In this work, we propose a robust approach to perform Anderson acceleration (AA) to speed up the training of SVM classifier models for multi-dataset training from the domain of biological sequencing. We regularize AA by including it in the loss optimization of simple linear classifier models (SVMs) and classical ML training, in contrast to previous work in complex deep learning models.
We numerically demonstrate the effectiveness of the proposed acceleration by comparing the training loss with an increasing number of iterations on different sets of biological sequences. The results show that using AA significantly improves convergence and efficiently accelerates the training of traditional ML models.

\section{Related Work}
Iterative optimization methods like gradient descent and its variants are widely used for training ML models, but convergence can be slow, especially for high-dimensional problems. Anderson acceleration (AA) is a technique for speeding up the convergence of these methods by exploiting the geometry of the search space. It was first introduced by Anderson~\cite{anderson1965iterative} as a way to accelerate the convergence of the conjugate gradient method and has since been applied to a variety of optimization techniques, such as Newton's method~\cite{an2017anderson}, stochastic gradient descent~\cite{mai2020anderson}, and the Nelder-Mead simplex algorithm~\cite{barton1991modifications}.

In recent years, there has been a growing interest in using Anderson acceleration for training deep neural networks, where it has been applied to a variety of tasks, such as image classification~\cite{pasini2022anderson}, natural language processing and reinforcement learning~\cite{zuo2022offline}. 
Anderson acceleration is particularly well-suited for deep learning problems, where it has been shown to improve convergence and generalization performance~\cite{pasini2021stable,bertrand2022beyond}. A method to estimate a sparse generalized linear model with convex or non-convex separable penalties using Anderson acceleration is also proposed in~\cite{bertrand2022beyond}.
In these approaches, Anderson acceleration has been shown to improve convergence and generalization performance compared to traditional optimization methods, such as gradient descent. In addition, it has also been applied to logistic regression~\cite{bertrand2021anderson} and other ML models.

However, despite these advances, Anderson acceleration has not been widely applied to classical machine learning classifiers, such as support vector machines (SVM), despite the potential for improved convergence rates. This is particularly relevant for tabular data, where classical machine learning classifiers are widely used. The limited exploration of Anderson acceleration in classical machine learning classifiers is surprising, given that the technique is effective in improving convergence in other optimization problems~\cite{anderson1965iterative,an2017anderson,mai2020anderson,barton1991modifications}.

Another area where Anderson acceleration has shown promise is in training sparse models, such as sparse coding and dictionary learning~\cite{rodriguez2021computational}. In these applications, Anderson acceleration effectively improves convergence and achieves sparsity, an essential consideration in many machine-learning models.

In recent years, researchers have also explored the use of Anderson acceleration in the training of generative adversarial networks (GANs)~\cite{he2022solve}. In these applications, Anderson acceleration has been shown to improve convergence and stability and to produce high-quality synthesized images.

Finally, it's worth noting that Anderson acceleration has also been applied to the training of robust models that are robust to outlier examples and to adversarial attacks~\cite{garstka2022safeguarded}. In these applications, Anderson acceleration effectively improves the robustness of machine learning models and defends against adversarial attacks.

\section{Proposed Approach}
This section first discusses the algorithm we use for the proposed method. Later, we discuss the theoretical understanding of Anderson Acceleration and the assumptions considered.

Anderson Acceleration (AA) attempts to make greater use of previous data than the fixed-point iteration, which only takes the most recent iteration to produce a new estimate, $y_{ k+1} = g(y_k)$. The proposed method's algorithmic pseudocode is provided in Algorithm~\ref{algo_svm_anderson}, and the model training flow chart is shown in Figure~\ref{AA_flow_chart}. For model training, given the feature embedding X (or $\phi$) made from SARS-CoV-2 sequences and its lineage (variants) as labels Y, the first step involves the embedding generation using the methods discussed in Section~\ref{Section_Embedding_Methods}, the feature vector generated and the labels for the sequences are then supplied to the algorithm. In the algorithm, firstly, the weight vector is initialized with random values (1 $\times$ length of sequence). We then initialize the $iterLoss$ and $gradient$ values (lines 2-3 in Algorithm~\ref{algo_svm_anderson}) for each iteration. Afterward, for each input sample X and its label y, we predict using weight vector $\Vec{w}$ (line 5 in Algorithm~\ref{algo_svm_anderson}), the predicted value is normalized, and the gradient is updated (lines 5 and 6 in Algorithm~\ref{algo_svm_anderson}). Sample loss is updated, and the iteration loss list $iterLoss$ is maintained (lines 8 to 9 in Algorithm~\ref{algo_svm_anderson}). After every sample is processed, the gradient is averaged out, and weight history is maintained for the iteration (lines 11 and 12 in Algorithm~\ref{algo_svm_anderson}), also shown in Figure~\ref{AA_flow_chart}-e. Anderson acceleration is used to update the weight vector from the third iteration since we need at least two weight histories. The difference between the last two weight histories is computed and is multiplied with Anderson factor $\alpha$ as shown in lines 14 and 15 in Algorithm~\ref{algo_svm_anderson}, also shown in Figure Figure~\ref{AA_flow_chart}-ii. The loss and accuracy for the iteration are saved, and the next iteration is performed to do the same steps. Finally, after all iterations, the loss list is returned for the given input feature vectors. The loss for each Iteration is captured and argued to be the better option for faster convergence using Anderson Acceleration. 

\subsection{Anderson Acceleration}
One way to formally prove the convergence of Anderson acceleration is to use the concept of ``linear convergence'', which refers to the rate at which the optimization process approaches the optimal solution. Specifically, we can show that under certain conditions, the Anderson acceleration optimization process converges linearly, meaning that the error decreases by a constant factor at each iteration. This contrasts standard gradient descent, which converges at a sublinear rate (e.g., the error decreases by a factor less than 1 at each iteration).

To prove this result, we can start by considering the optimization problem in the form of a series of updates to the weight vector, where the update at each iteration is given by:

\begin{equation}
    w_{t+1} = w_t - \alpha \nabla f(w_t)
\end{equation}

where $w_t$ is weight vector at iteration $t$, $\alpha$ is the learning rate, and $\nabla f(w_t)$ is the gradient of the objective function $f$ at $w_t$. Now, we can add the Anderson acceleration term to the update, resulting in:

\begin{equation}
    w_{t+1} = w_t + \alpha (w_t - w_{t-1}) - \alpha \nabla f(w_t)
\end{equation}

Next, we can define the error at each iteration as:

\begin{equation}
    e_t = w_t - w^*
\end{equation}

where $w^*$ is the optimal weight vector. Now, we can substitute the expression for the update into the expression for the error and rearrange it to get:

\begin{equation}
    e_{t+1} = (1 - \alpha) e_t + \alpha e_{t-1} - \alpha \nabla f(w_t)
\end{equation}

where we have used the fact that $w^* = w_t - \nabla f(w_t)$. Now, we can define the ``damping factor'' as:

\begin{equation}
    \delta = 1 - \alpha
\end{equation}

and rewrite the expression for the error as:

\begin{equation}
    e_{t+1} = \delta e_t + (1 - \delta) e_{t-1}
\end{equation}

This expression has the form of a weighted average, where the weight of the current error is given by $\delta$, and the weight of the previous error is given by $1 - \delta$. Now, we can make the following assumptions:

\begin{enumerate}
    \item The objective function $f$ is continuously differentiable and has a Lipschitz continuous gradient, i.e., there exists a constant $L$ such that 
    \begin{equation}
        | \nabla f(x) - \nabla f(y) | \leq L | x - y |
    \end{equation}
    for all $x, y$.
    \item The objective function $f$ is bounded below, i.e., there exists a constant $f_{\min}$ such that $f(x) \geq f_{\min}$ for all $x$.
    \item The optimization algorithm is using a fixed step size $\alpha$, and the sequence of points ${x_k}$ generated by the algorithm satisfies
    \begin{equation}
        | x_{k+1} - x_k | \leq R
    \end{equation}
    for some constant $R$ and all $k$.
\end{enumerate}

These assumptions are typically made in the analysis of gradient descent algorithms. They allow us to establish certain convergence properties of the algorithm. Specifically, under these assumptions, it can be shown that the sequence of points generated by gradient descent with Anderson acceleration converges to a stationary point (a point where the gradient is zero) of the objective function $f$ at a rate of $O(1/k)$, where $k$ is the iteration number. This convergence rate is faster than $O(1/k^2)$ achieved by plain gradient descent without Anderson acceleration.

Intuitively, Anderson acceleration can be thought of as a way to incorporate information from past iterations into the current iteration to improve the convergence rate of the optimization algorithm. This is achieved using a weighted combination of the current gradient and the difference between the current and previous iterates. The weights are chosen such that the resulting update direction better approximates the true gradient at the current iterate, leading to faster convergence.

To compute the loss, we use ``cross-entropy loss" using the following expression:
\begin{equation}
    \text{Cross Entropy Loss} = -sum(y \times log(yPred + 1e-10))
\end{equation}
where y is the true label, and yPred is the predicted label. The $1e-10$ is added to the yPred to avoid the log of zero, which will cause an infinity error. The negative sign ensures the optimization problem is formulated as a minimization problem (hence, our loss can be negative).

\begin{figure}[h!]
  \centering
  \includegraphics[scale=0.2]{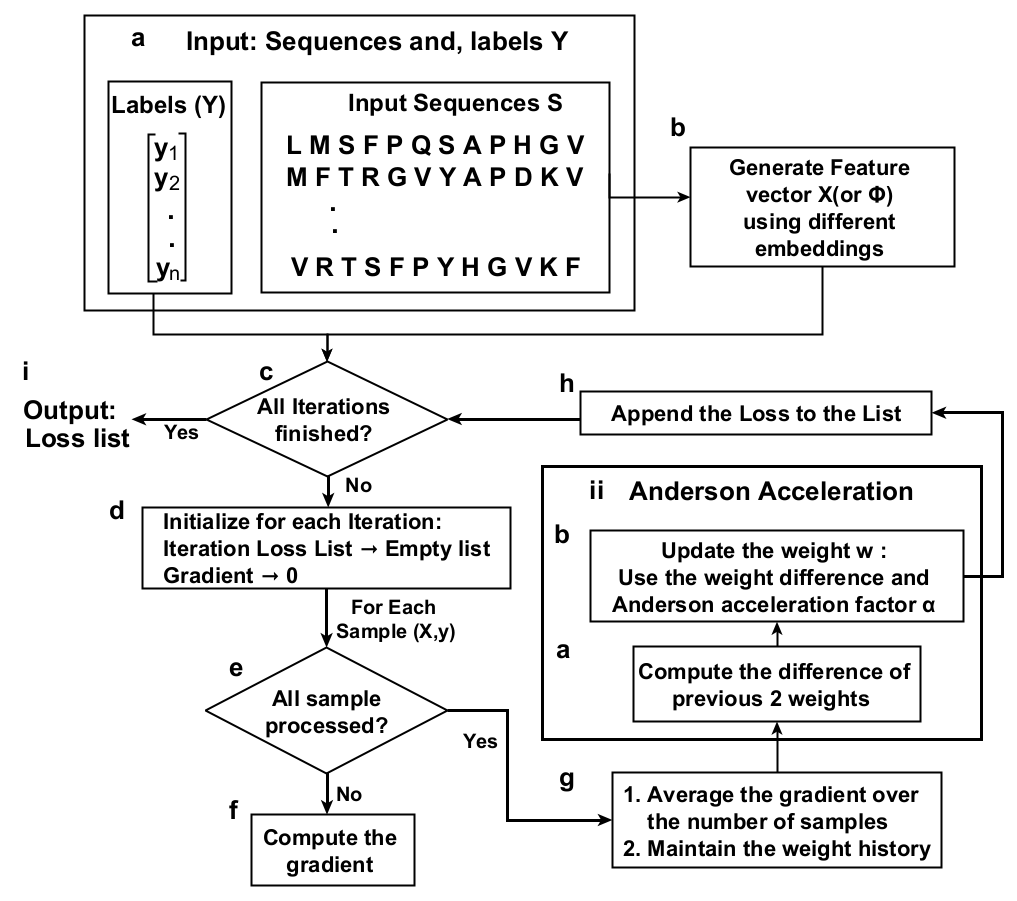}
  \caption{Flow chart for Anderson Acceleration training. Given the input data (X,y), we
    First, predict with initial weight w; this w is updated using SGD and modified with AA in consecutive steps to compute the loss and accuracy for the prediction. 
    }
  \label{AA_flow_chart}
\end{figure}

The flowchart for training with Anderson Acceleration(AA) is shown in Figure~\ref{AA_flow_chart}. We provide the Feature vectors X (or $\phi$) as input along with the labels Y. Few parameter initializations are required, such as the number of iterations and the weight vector initialized with random values. Anderson acceleration factor $\alpha$, for which we tried several values to study its impact and select the best value. An empty list for loss is also shown in Figure~\ref{AA_flow_chart}-b. For a given number of iterations, we process the samples to compute the gradient and loss for the sample. The gradient is averaged out, and we update the weight using Anderson Acceleration for that iteration, also shown in Figure~\ref{AA_flow_chart}-ii. The process is repeated for the given number of iterations.

\begin{algorithm}[h!]
\caption{SVM With Anderson Acceleration 
}
\label{algo_svm_anderson}
\begin{algorithmic}[1]
\scriptsize
\Statex \textbf{Input:} Embedded training data $X$, target labels $y$, no. of iterations $iter$, Anderson acceleration factor $\alpha$, 
initial weight vector $\Vec{w} \gets \text{randn}(len(X[0])$, and loss list $loss$
\Statex \textbf{Initialization:} 
\Statex $\Vec{w} \gets \text{randn}(X.\text{shape}[1])$  \Comment{Initial random weight}
\Statex $loss \gets []$  \Comment{Empty list} 
\Statex  $wHistory \gets \Vec{[w]}$ \Comment{Weight vector history}
\For{$i \in \{1, 2, \ldots, iter\}$}
\State  $iterLoss \gets 0$ \Comment{Initial loss}
\State $grad \gets \textbf{0}$ \Comment{Gradient}
\For{$(x, y) \in (X, y)$}
\State$yPred \gets w \cdot x$ \Comment{Predict the class
}
\State $yPred \gets \frac{yPred} {\text{sum}(yPred)}$ \Comment{Normalize the prediction}

\State $grad \gets grad + y - yPred$ 
\State $sampleLoss -= \text{sum}(y \times \text{log}(yPred))$

\State $iterLoss += sampleLoss$ 
\EndFor
\State $grad \gets \frac{grad}{\text{len}(X)}$ \Comment{Avg the grad over total samples}
\State $wHistory += \Vec{[w]}$ \Comment{Add updated $\Vec{w}$ to history}
\If{$\text{len}(wHistory) > 2$}
\State $diff \gets wHistory[i-1] - wHistory[i-2]$ 
\State $\Vec{w} \gets \Vec{w} + \alpha \times diff + grad$ \Comment{Update $\vec{w}$ using AA} 
\Else
\State $\Vec{w} \gets \Vec{w} + grad$ \Comment{Use gradient to update $\vec{w}$}
\EndIf
\State $loss.append([\frac{iterLoss} {\text{len}(X)}])$ \Comment{Append loss}
\EndFor
\State \textbf{return} $loss$
	\end{algorithmic}
\end{algorithm}

\subsection{Embedding Methods}
\label{Section_Embedding_Methods}
We employ the three representation learning techniques below to convert the biological sequences into low-dimensional embeddings.

\subsubsection{Spike2Vec~\cite{ali2021spike2vec}}
This technique offers numerical embedding of the supplied input spike sequences to facilitate the use of ML models. Initially, it produces $k$-mers
of the supplied spike sequence because $k$-mers are known to maintain the sequence's ordering information. For a sequence of length $N$, the total number of $k$-mers produced is $N - k + 1$. 
For every particular sequence, $k$-mers is a collection of (contiguous) amino acids (also known as mers) of length $k$. (also called nGram in the NLP domain).
To convert the $k$-mers alphabetical data into a numerical representation, the Spike2Vec computes the frequency vector based on $k$-mers. This vector comprises the counts of each $k$-mer in the sequence. A fixed-length feature vector is then made using the generated $k$-mers and their frequencies in a sequence. The character alphabet $\Sigma$ and the length of the $k$-mers are used to calculate the length of this feature vector, which is $|\Sigma|^k$.

\subsubsection{Minimizer~\cite{robertsReducingStorageRequirements2004a}}
A minimizer, also known as an $m$-mer, is the substring of consecutive letters (amino acids) of length $m$ from a given $k$-mer that is lexicographically the smallest in both forward and backward order of the $k$-mer, where $m<k$ and is fixed. 
The repetition of $k$-mers in a long sequence, which increases computing and storage costs, is one of the fundamental issues with $k$-mers. Minimizers could be used to get rid of this redundant information. 
The lexicographically lowest $m$-mer from both the forward and backward $k$-mers is then used to calculate a minimizer. Two parameters ($k, $m) are provided to the minimizer. The $k$-mer's length is $k$, where $k=9$ in our example, and the $m$-mer's size is $m$, where $m=3$. It extracts $k$-mers of a sequence given to it. Then, it calculates a corresponding $m$-mer for each $k$-mer (minimizer). A fixed-length feature vector is then made using the generated minimizers $m$-mers and their frequencies in a sequence. The character alphabet $\Sigma$ and the length of the $m$-mers are used to calculate the length of this feature vector, which is $|\Sigma|^m$.

\subsubsection{Spaced k-mer~\cite{singh2017gakco}}
The performance of sequence classification is significantly impacted by the size and sparsity of feature vectors for sequences based on $k$-mers frequencies. The idea of employing non-contiguous length $k$ sub-sequences ($g$-mers), proposed by spaced $k$-mers, to create compact feature vectors with reduced sparsity and size. It first computed $g$-mers using a spike sequence as input. We calculate $k$-mers, where $kg$, from those $g$-mers. To conduct the trials, we used $k=4$ and $g=9$. The gap's dimensions are determined by $g-k$. However, this approach still involves bin scanning, which is computationally expensive and generates very high dimensional feature representation.
We took 500 Principle components by applying PCA~\cite{wold1987principal1} for high dimensional embeddings (feature vector length $> 1000$).

\section{Experimental Evaluation}
To perform evaluation, we use datasets including Genome and Host. The details are as follows:

\subsection{Dataset Statistics}
\subsubsection{Genome Dataset}
Using the well-known and widely used database of SARS-CoV-2, GISAID~\cite{gisaid_website_url}, we retrieve the full-length nucleotide sequences of the coronavirus. Our dataset includes the COVID-19 variant information and $8220$ nucleotide sequences. In our sample, there are 41 different Lineages altogether. The goal is to classify the sequences and predict the Lineage it belongs to.

\subsubsection{Host Dataset}
The National Institute of Allergy and Infectious Disease (NIAID) Virus Pathogen Database, Investigation Resource
(ViPR)~\cite{pickett2012vipr}, and GISAID
was used to retrieve the Spike protein sequences from a collection of spike sequences from several clades of the Coronaviridae family, along with details about the hosts that each spike sequence has infected. The hostname is used as the class label in our classification tasks for this dataset. It displays the distribution of the dataset across the various host types (grouped by family).

\subsection{Evaluation Metrics}
For performance evaluation of SVM without and with Anderson acceleration, we use cross-entropy loss.
The cross-entropy loss, also known as the negative log-likelihood loss, is commonly used in supervised learning problems with categorical targets. The cross-entropy loss for a single sample can be expressed mathematically as follows: $
L = -\log \left(\frac{e^{f_{y_i}}}{\sum_{j=1}^{K} e^{f_j}}\right)$, where $f_{y_i}$ is the predicted score for the correct class and $K$ is the number of classes. The cross-entropy loss is averaged over the entire training set to obtain the final objective function optimized during training.

The cross-entropy loss penalizes the predicted scores for the incorrect classes and rewards the predicted score for the correct class. During training, the goal is to minimize the cross-entropy loss so that the predicted scores for the correct class are as high as possible compared to those for incorrect classes.

\section{Results And Discussion}
In this section, we report results comparison without and with Anderson acceleration using cross-entropy loss for different biological sequence datasets.

\subsection{Results For Genome Data}
\begin{figure}[h!]
  \centering
   \begin{subfigure}{.15\textwidth}
  \centering
  \includegraphics[scale=0.32]{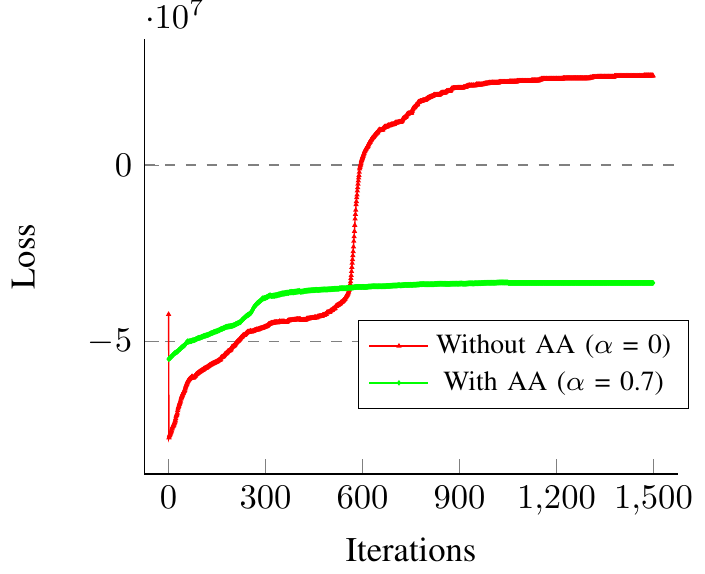}
  \caption{Spike2Vec}
  \label{}
  \end{subfigure}%
   \begin{subfigure}{.15\textwidth}
  \centering
  \includegraphics[scale=0.32]{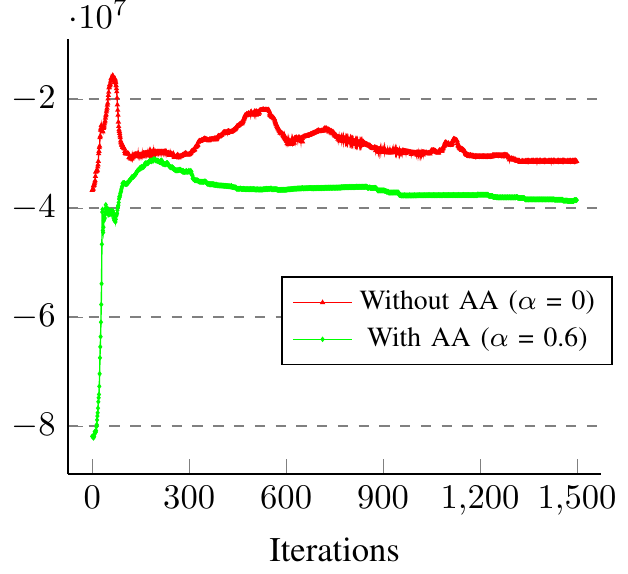}
  \caption{Minimizer}
  \label{}
  \end{subfigure}%
   \begin{subfigure}{.15\textwidth}
  \centering
  \includegraphics[scale=0.32]{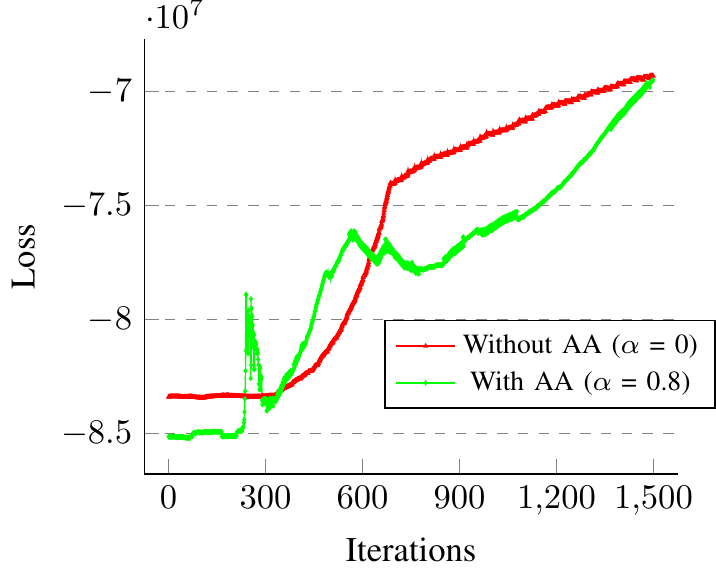}
  \caption{Spaced Kmer}
  \label{}
  \end{subfigure}
  \caption{Loss plots with AA (green line) and without AA (red line) for \textbf{Genome data}. The x-axis shows the increasing number of iterations, while the y-axis shows the cross entropy loss.  The figure is best seen in color.
  }
 \label{fig_losses_Genome}
\end{figure}

The results for genome data using all embedding methods are reported in Figure~\ref{fig_losses_Genome} for the best value of Anderson Acceleration (AA) factor $\alpha$. We use cross-validation to get the best value for $\alpha$ ranging from (0, 0.1, 0.2, $\cdots$, 1.0) for respective embeddings, where 0 implies no AA and 1.0 shows maximum AA.
For Spike2Vec embedding, we can observe that although cross-entropy loss without Anderson acceleration is smaller with fewer iterations, as we increase the iterations, the loss increases too. On the other hand, the loss does not increase significantly while using Anderson acceleration in SVM. Moreover, with AA, the loss started to converge after $300$ iterations, which is almost half compared to the loss convergence without AA (i.e., $\approx$ 600 iterations). For Minimizer-based embedding, although we can observe more fluctuation in loss compared to Spike2Vec, the loss (and convergence) is less when SVM is used along with AA. Similarly, the behavior of spaced $k$-mers-based embedding differs from both Spike2Vec and Minimizer-based embedding. Although we can see an overall increasing trend in loss with an increasing number of iterations, the SVM with AA loss is lower than without AA when the number of iterations increases. Overall, it is evident from all three embedding results that the loss with AA is less than the loss without AA for different embedding methods as we increase the number of iterations, showing the significance of using AA for the training of SVM.

\subsection{Results For Host Data}
\begin{figure}[h!]
  \centering
   \begin{subfigure}{.16\textwidth}
  \centering
  \includegraphics[scale=0.35]{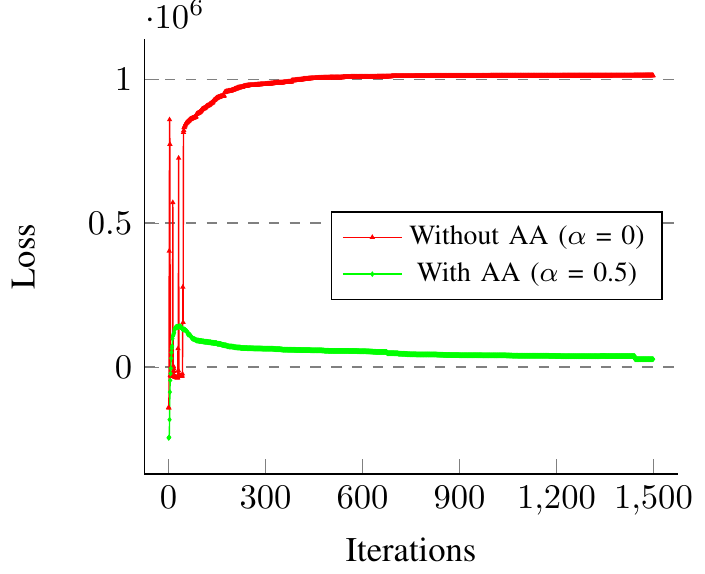}
  \caption{Spike2Vec}
  \label{}
  \end{subfigure}%
   \begin{subfigure}{.16\textwidth}
  \centering
  \includegraphics[scale=0.35]{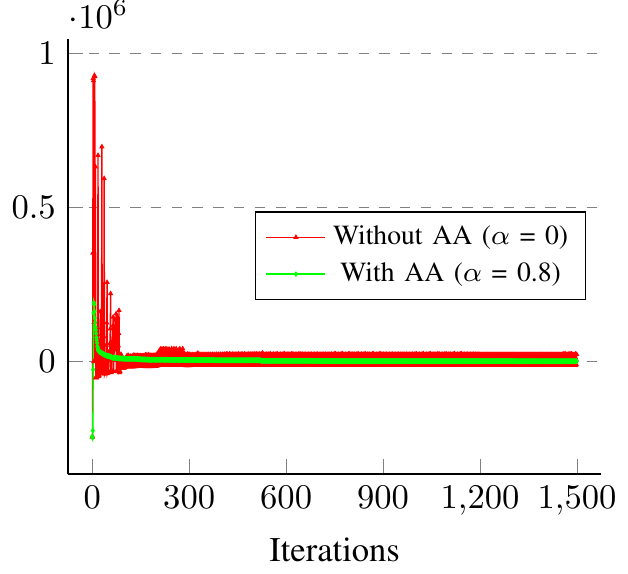}
  \caption{Minimizer}
  \label{}
  \end{subfigure}%
   \begin{subfigure}{.16\textwidth}
  \centering
  \includegraphics[scale=0.35]{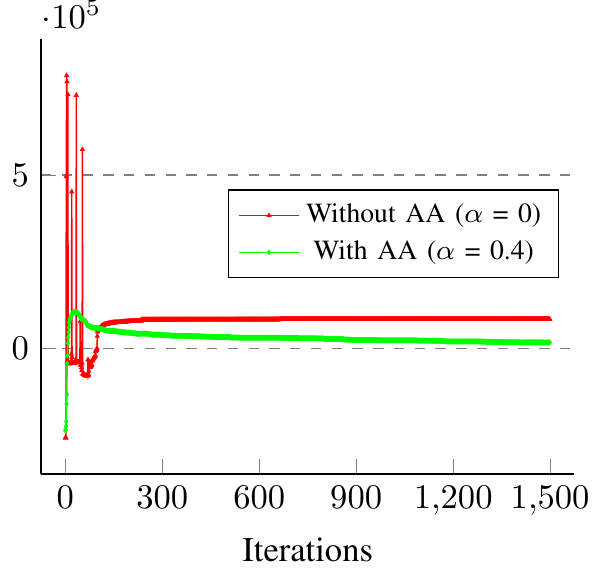}
  \caption{Spaced Kmer}
  \label{}
  \end{subfigure}
  \caption{Loss plots with AA (green line) and without AA (red line) for \textbf{Host data}. The x-axis shows the increasing number of iterations, while the y-axis shows the cross entropy loss.  The figure is best seen in color.
  }
 \label{fig_losses_Host}
 
\end{figure}

The results for host data using all embedding methods are reported in Figure~\ref{fig_losses_Host} for the best value of Anderson Acceleration (AA) factor $\alpha$. We use cross-validation to get the best value for $\alpha$ ranging from (0, 0.1, 0.2, $\cdots$, 1.0) for respective embeddings, where 0 implies no AA and 1.0 shows maximum AA.
For Spike2Vec-based embedding, the behavior is not different from the same embedding in the case of Genome data. Although SVM without and with Anderson acceleration converges very fast (i.e., in $<100$ iterations), the cross entropy loss with AA is smaller than SVM without AA. We observed some improvement in the SVM without AA in the Minimizer and Spaced $k$-mers-based embedding methods. However, when the number of iterations is smaller, we can observe some fluctuation in the cross-entropy loss for SVM without AA, compared to the smooth loss curve for SVM with AA, showing its significance in efficient training of the SVM classifier.

Overall, we can observe from the results of all two datasets and all three embeddings that SVM with Anderson acceleration converges faster and shows less cross-entropy loss than SVM without Anderson acceleration. This behavior shows the effectiveness of SVM when used with Anderson acceleration in terms of training time and performance when applied in a real-world setting (as we show its performance on a $3$ real-world set of biological sequences). We believe this study can open more opportunities for researchers to explore further the applications of Anderson acceleration for other datasets (e.g., from other domains such as finance and images) when applied to other machine learning models such as perceptron. These future studies can help us further understand and improve the predictive performance of classical ML models when applied in a real-world (biological) setting.

\section{Conclusion}
In summary, this study provides a novel use of Anderson acceleration in the bioinformatics area of support vector machine (SVM) classifier. Our experiments on several sequence-based bioinformatics datasets show that Anderson acceleration results in a considerable decrease in training loss and improved convergence compared to the standard SVM. 
In the future, we will investigate more traditional linear classifier models, such as the Perceptron, and bigger biological data to assess their scalability and resilience. Moreover, evaluating the robustness and generalizability of the proposed Anderson acceleration method is also an interesting future extension.

\bibliography{references}

\end{document}